\documentclass[12pt,a4paper,notitlepage]{article}
\usepackage{times}
\usepackage{latexsym}
\usepackage{makecell}
\usepackage{graphics}
\usepackage{CJKutf8}
\usepackage[bottom]{footmisc}
\usepackage{ulem}
\usepackage[T1]{fontenc}
\usepackage[utf8]{inputenc}
\usepackage{hyperref}
\usepackage{microtype}
\usepackage{graphicx}
\usepackage{caption}
\usepackage{amsmath}
\usepackage{mathtools}
\usepackage[toc]{appendix}
\usepackage{multirow}
\usepackage{amsfonts}
\usepackage{xcolor}
\usepackage{booktabs}
\usepackage{url}
\usepackage{listings}
\usepackage{siunitx}
\usepackage{pgfplotstable}
\usepackage{tikz}
\usetikzlibrary{fit}
\usepackage{blkarray}
\usepackage{enumitem}
\usepackage[round]{natbib}
\usepackage{setspace}
\usepackage{sectsty}
\usepackage{url}
\sectionfont{\color{blue}}
\subsectionfont{\color{red}}
\hypersetup{colorlinks,linkcolor={blue},citecolor={blue},urlcolor={red}} 

\title{A Novel Dependency Framework for Enhancing Discourse Data Analysis}
\author{
	Kun Sun \\
	University of Tübingen \\
	\texttt{kun.sun@uni-tuebingen.de}
	\and
	Rong Wang \\
	University of Stuttgart \\
	\texttt{rongw.de@gmail.com}
}
\date{}

\begin{document}
\maketitle

\begin{abstract}

The development of different theories of discourse structure has led to the establishment of discourse corpora based on these theories. However, the existence of discourse corpora established on different theoretical bases creates challenges when it comes to exploring them in a consistent and cohesive way. This study has as its primary focus the conversion of PDTB annotations into dependency structures. It employs refined BERT-based discourse parsers to test the validity of the dependency data derived from the PDTB-style corpora in English, Chinese, and several other languages. By converting both PDTB and RST annotations for the same texts into dependencies, this study also applies ``dependency distance'' metrics to examine the correlation between RST dependencies and PDTB dependencies in English. The results show that the PDTB dependency data is valid and that there is a strong correlation between the two types of dependency distance. This study presents a comprehensive approach for analyzing and evaluating discourse corpora by employing discourse dependencies to achieve unified analysis. By applying dependency representations, we can extract data from PDTB, RST, and SDRT corpora in a coherent and unified manner. Moreover, the cross-linguistic validation establishes the framework's generalizability beyond English. The establishment of this comprehensive dependency framework overcomes limitations of existing discourse corpora, supporting a diverse range of algorithms and facilitating further studies in computational discourse analysis and language sciences.

	\end{abstract}


	\section{Introduction}
	
	Discourse or text generally has multiple clauses or sentences. The parts of discourse are interrelated and form a coherent whole that clearly expresses a meaning. 
	Discourse in the wider sense underlies disciplines such as law, religion, politics, science amongst others. \textit{Discourse structure}, like syntax, concerns the ways in which discourse units are brought together to form a coherent discourse. Discourse structure mostly concerns the logical and semantic interrelations of discourse units (or elementary discourse units, EDUs). 
	\textit{Discourse relation} is the semantic or logical meaning of the connections between discourse units. It is a central concern in discourse structure research. The structure of discourse has already been extensively investigated from theoretical, experimental and computational perspectives. The processing of discourse structure (e.g., discourse parsing) is of significant importance to linguistic research, particularly in the field of Natural Language Processing (NLP)  \citep{hou2020rhetorical}. 
	
	Researchers have formulated numerous theories that interpret discourse structure, showcasing a wide range of perspectives on the interplay between discourse units and textual coherence. For example, RST (Rhetorical Structure Theory, \citet{mann1988rhetorical}) is one influential approach in this field \cite{taboada2006rhetorical}. It describes textual coherence using the rhetorical relations between EDUs and postulates a hierarchical tree structure. 
	By contrast, D-LTAG theory (a lexicalized Tree Adjoining Grammar for discourse) holds that discourse relations can be lexicalized \cite{webber2004d}. Based on the fact that discourse connectives signal discourse relations (e.g., ``because, although, when''), this theory treats two discourse units as linked by a connective, which means that one discourse unit and the connective together constitute a dependency relation. Relations through discourse connectives can be treated as local constituency structure by discourse units in the PDTB. Another influential theory, SDRT (Segmented Discourse Representation Theory, \citet{asher2003logics}), combines dynamic semantics with a discourse structure defined via rhetorical relations between segments. These approaches to discourse structure thus reflect their different areas of focus.

	
	Not only have many different theories of discourse structure been developed but discourse corpora have also been established based on these theories (\citealp{webber2019penn}; \citealp{carlson2003building}; \citealp{abzianidze2017parallel}). In light of their great influence, these discourse corpora styles have been annotated in a number of languages. However, this creates a problem for researchers: The very fact these corpora have been established on different theoretical bases makes it difficult to explore discourse corpora in a consistent and unified way. 
	Due to the aforementioned theoretical distinctions that underlie these ways of annotating discourse corpora, it is difficult to explore different discourse corpora in a consistent fashion despite the fact that it is possible to find some mappings between them \cite{demberg2017compatible} and despite the fact that some attempts have been made to annotate different styles of discourse structure using the same texts \cite{stede2016parallel}. These attempts did not, however, discover a more general structure that can be used to represent discourse. This makes applying these approaches more generally difficult. This means that it is hard to apply these approaches more generally. 
	
	In order to deal with these difficulties, we want to find a more general structure or framework for representing the available discourse corpora and so enable more algorithms to process data. We also need a framework for describing and explaining discourse structure quantitatively. The establishment of a comprehensive framework or structure can allow a diverse range of well-developed algorithms to explore unified data and this in turn can facilitate further studies in the field. Recent studies have shown that a rhetorical structure can be converted into dependency representations (\citealp{yoshida2014dependency}; \citealp{li2014text}). SDRT relations have been studied with respect to the conversion of dependency relations \cite{danlos2005comparing}. We suppose that PDTB relations can also be converted into dependency representations. All this indicates that we can employ dependency representations to generalize discourse relations in different discourse corpora if we seek their largest common denominator, and apply these discourse theories more widely.
	
	The primary objective of the current study is to address the challenges posed by the existence of multiple discourse corpora based on different theoretical frameworks. Specifically, the present study aims to: 1) Convert PDTB Annotations into Dependency Structures: Develop a method to transform the PDTB annotations into dependency representations, thereby enabling a more unified approach to discourse analysis. 2) Validate Dependency Data: Employ updated BERT-based discourse parsers to test the validity of the dependency data derived from PDTB-style corpora across multiple languages. 3) Examine Correlations: Analyze the correlation between RST dependencies and PDTB dependencies using some metrics.

	\section{Related Work}
	
Previous theories and corpora reflect the many and different perspectives on textual coherence. This section describes the key characteristics on the discourse corpora followed by RST, D-LTAG, and SDRT.
	
\begin{figure}
		\includegraphics[width=0.96\textwidth]{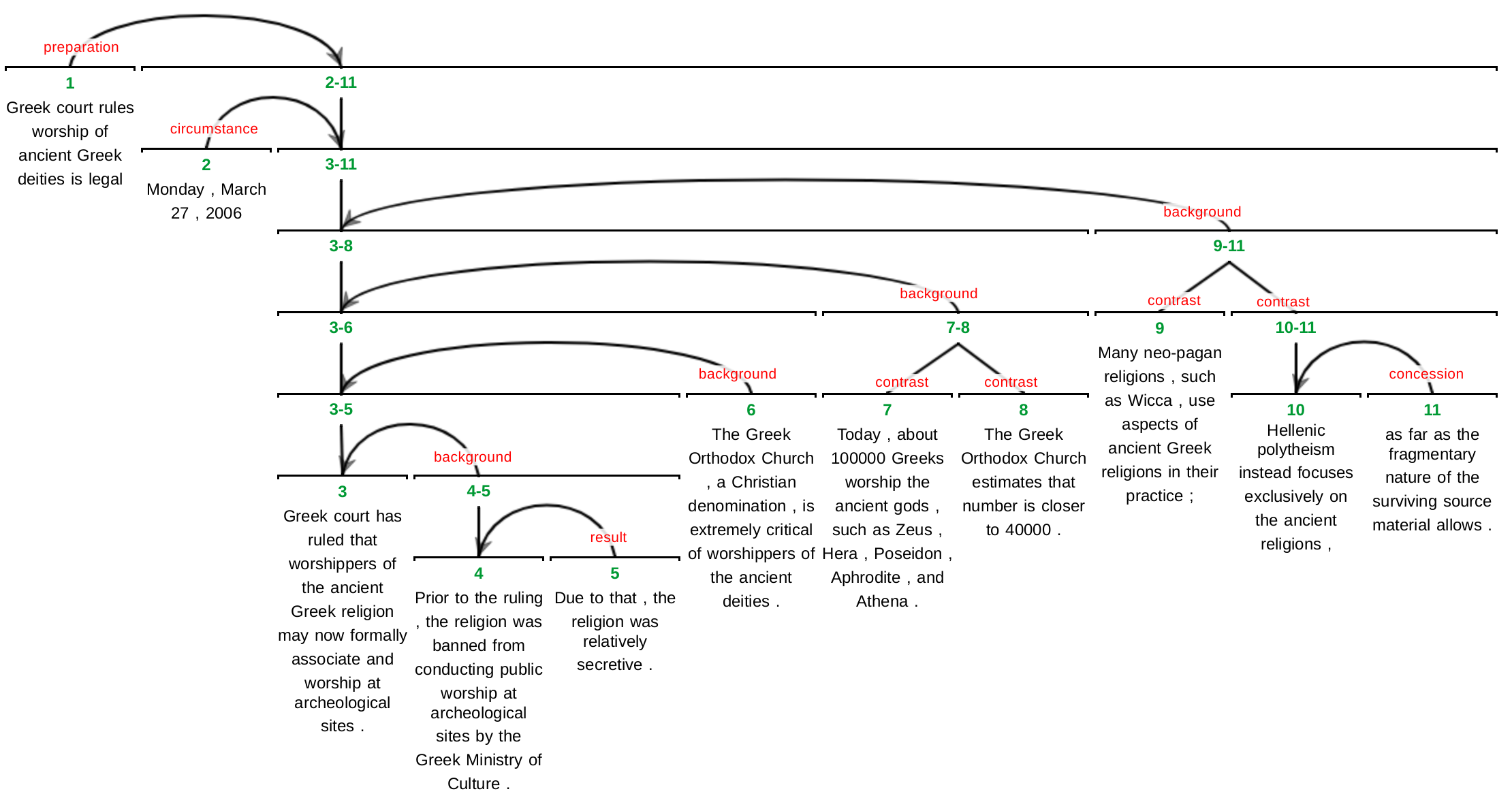}
		\caption{The RST-tree of an example}
		\label{fig:1}	
	\end{figure}

RST is an influential approach in discourse structure studies that describes textual coherence using the rhetorical relations between EDUs and postulates a hierarchical tree structure. 
	The RST is a constituency-based theory, which means that discourse units combine with discourse relations to form recursively larger units and these ultimately form a global document. RST corpora (such as the RST discourse treebank (RST-DT), \citet{carlson2003building}) help enormously in analyzing discourse structure quantitatively and in the automatic processing of texts. 
	What is characteristic of RST annotations is that each discourse unit has to be related to the overall discourse structure and represented using a (rhetorical) tree structure. A second essential characteristic of the RST-DT is the assignment of nuclearity: EDUs  are characterized as nuclei or satellites. The nucleus is the most central part of a relation in the text (the node where the arrow point is located is the nucleus in Fig.\ref{fig:1}),  while the satellites support the nucleus. A discourse relation can be either mono-nuclear (e.g.,4-5 in Fig.\ref{fig:1}) or multi-nuclear (e.g.,7-8 in Fig.\ref{fig:1}). All RST relation types can be classified into one of 16 classes. For instance, ``cause'', ``result'' and ``consequence'' belong to the class of ``CAUSE''.
	
	The other influential discourse corpus, the Penn Discourse Treebank (PDTB 3.0, \citet{webber2019penn}), annotates English discourse structures following the D-LTAG. But because the D-LTAG looks at each relation individually and disregards most of the surrounding structures, it does not take into account the global structure of the text. This is the approach used in shallow discourse parsing. The PDTB uses a discourse connective as a head for forming \textit{dependency} concerning local discourse structure. The PDTB3.0 distinguishes 43 relation labels with annotations. These labels are organized in a hierarchy consisting of three levels (i.e., sense, class, type). To sum up, RST uses a rhetorical relation tree to describe global discourse structure, while D-LTAG (i.e., PDTB) uses shallow discourse parsing. Moreover, several discourse corpora have been established following the SDRT, such as the Parallel Meaning Bank \citep{abzianidze2017parallel}, and the STAC corpus \citep{asher2016discourse}. 
	
	As mentioned in introduction, it is possible to find some mappings between them. Further, attempts have been made to annotate different styles of discourse structure using the same texts \cite{stede2016parallel}. The methods of mapping the discourse relations of different frameworks were presented by \cite{scheffler2016adding}. The main approach is to use the texts that were annotated under different frameworks. \citet{scheffler2016adding} focus on mapping between explicit PDTB discourse connectives and RST rhetorical relations by using the Potsdam Commentary Corpus \cite{stede2014potsdam}, which contains both RST and PDTB annotations in German. 
	\citet{demberg2017compatible} found that 76\% RST relations are mapped with PDTB ones in English, which is similar to the finding of \citet{scheffler2016adding} in German. \citet{sanders2021unifying} formulated a set of core relations that are shared by existing frameworks but are open and extensible in use, with the outcome being ISO-DR-Core \citep{bunt2016iso}. However, these attempts did not find a more general (data) structure that can be used to represent different types of discourse structure. 
	
	There is a lack of consensus in a number of studies with respect to how to represent discourse structure. Yet nonetheless, we have found that several types of discourse relations can be converted into discourse dependency. For instance, there has been some successes in converting RST relations into dependency relations, as detailed in numerous studies (\citealp{hirao2013single}; \citealp{venant2013expressivity}; \citealp{li2014text}). SDRT relations have also been converted into dependency (\citealp{danlos2005comparing}; \citealp{stede2016parallel}). 
	
	Inspired by recent studies, 
	we convert PDTB relations into dependency data. 
	The discourse dependency structure can be taken as a common structure that the other structures can be converted into, given that PDTB can also be converted into dependency structure. In this way, we can use dependency as a unified framework for describing and explaining discourse structure formally, computationally and quantitatively. The dependency approach has been explored over a long period of time in linguistics, and dependency parsing has been widely used in computational linguistics (\citealp{liu2017dependency}; \citealp{gibson2019efficiency}; \citealp{de2021universal}). 
	  Dependency-based annotation has been adopted to establish discourse corpora.(\citealp{yang2018scidtb}; \citealp{nishida2022out}).


	\section{Methods}

	\subsection{Dependency parsing}
	\subsubsection{PDTB converted into dependency representations}
	
	A considerable amount of studies have successfully explored the conversion of RST- style and SDRT-style annotations into dependency representations. Currently two algorithms were used in this undertaking (\citealp{hirao2013single}; \citealp{li2014text}). 
	We focus on how to convert PDTB relations into dependency representations.
	
	Using a discourse connective, a dependency between two arguments (i.e., two discourse units or EDUs) in the PDTB can be treated as the semantic relation between two EDUs in practice because people tend to be concerned with the relation between two EDUs in the PDTB. \citet{yi2021unifying} used this to convert Chinese PDTB relations into dependency tree. \citet{yi2021unifying} did not make use of the PDTB third-level annotation information to help in distinguishing the head and dependent in dependency. However, converting PDTB relations into a global dependency tree actually violates the original purpose of shallow or local parsing in the PDTB. That is why we will not convert PDTB structure into a dependency tree. We adopt the dependency structure to preserve the original PDTB information and characteristics to the maximum extent possible. 
	
	In most cases, two adjacent discourse units are connected by a simple semantic relation. However, PDTB can also form some complicated structures and these are very similar to the local (syntactic) constituency structure (\citealp{johansson2007extended}; \citealp{li2014text}). As is well known, an RST tree represents a global constituency structure. However, it is possible to convert an RST constituency tree into a dependency tree. Similarly, the local constituency structure of PDTB can also be transformed into a local dependency structure. 
	
	Dependency grammar holds that the grammatical relation of binary relations comprise these dependency structures, and an individual dependency relation consist of a \textit{head} and a \textit{dependent}(or head vs. subordinate, head vs. governor) (\citealp{tesniere2015elements}; \citealp{hudson1984word}). Distinguishing the head and dependent in a discourse relation is of crucial importance when dependency parsing is applied. Although the PDTB annotations do not provide us with explicit information on the head or dependent, we can still use the existing PDTB annotation information to distinguish which EDU is the head or dependent. The PDTB annotation system is a hierarchical and it has three-class annotations, that is, \texttt{sense, class and type} (see Fig. \ref{fig:pdtbann} in the Appendix A.2). From the annotation information on \textit{type}, we can obtain the knowledge that its corresponding annotations on sense and class are \textit{symmetric} or \textit{asymmetric}. For example, ``\texttt{\footnotesize{when|Contingency.Condition.Arg2-as-cond}}'' is an asymmetric discourse relation. The reason for this is that when the second argument (the second EDU) is a conditional clause, this indicates that the first discourse unit includes more important information, that is, the first discourse unit is the ``head'' and the second is ``dependent(subordinate)''. Evidently, the dependency between a connective and its arguments differs fundamentally from the dependency relation between two arguments (i.e., two discourse units). The concept of ``head'' in dependency refers to a discourse unit with the more important information. In contrast, the ``head'' is the discourse connective in PDTB. Although the identical term is used, they are distinct in their meanings. In contrast, some relations are symmetric. For example, when a discourse relation is annotated with ``similarity'', we suppose that two discourse units may be equally important. According to \citet{bunt2016iso} and \citet{webber2019penn}, we believe that when the third-class tags with ``\texttt{Arg-as-XX}'' could provide the information on ``asymmetric'' EDUs. The remaining annotations are regarded as symmetric. 
	We will augment our manuscript with additional sentences to highlight these distinctions. Among 22 classes in the PDTB 3.0., 10 classes annotations contain asymmetric information.  Table \ref{tab:asymmetric} is a summary of asymmetric tags used in PDTB3.0. Asymmetric annotations can provide us with adequate information on how to distinguish which discourse unit is the head or which one is dependent in a discourse relation. 
	
	\begin{table}
		\centering
		\scalebox{0.66}{
			\begin{tabular}{|l|l|l|l|}
				\hline
				Type & (PDTB) Sense & (PDTB) Class & (PDTB) Type \\ [0.5ex]
				\hline\hline 
				Asymmetric & CONTINGENCY & Condition  & \makecell{Arg2-as-cond (Arg2-``head'');\\arg2-as-cond (Arg1-``head'')}\\
				\hline 
				Asymmetric & CONTINGENCY & negative-condition & \makecell{Arg1-as-negcond (Arg2-``head'');\\arg2-as-negcond (Arg1-``head'')} \\
				\hline 
				Asymmetric & CONTINGENCY &  purpose & \makecell{Arg1-as-goal (Arg2-``head'');\\arg2-as-goal (Arg1-``head'')} \\
				\hline 
				Asymmetric & COMPARISON & Concession & \makecell{Arg1-as-denier (Arg2-``head'');\\arg2-as-denier (Arg1-``head'')} \\	
				\hline 	
				Asymmetric & COMPARISON & Exception & \makecell{Arg1-as-except (Arg2-``head'');\\arg2-as-except (Arg1-``head'')} \\
				\hline 
				Asymmetric & COMPARISON & Instantiation & \makecell{Arg1-as-instance (Arg2-``head'');\\arg2-as-instance (Arg1-``head'')} \\
				\hline 
				Asymmetric & COMPARISON & Level-of-detail & \makecell{Arg1-as-detail (Arg2-``head'');\\arg2-as-detail (Arg1-``head'')}\\
				\hline 
				Asymmetric & COMPARISON & Manner & \makecell{Arg1-as-manner (Arg2-``head'');\\arg2-as-manner (Arg1-``head'')} \\
				\hline 
				Asymmetric & COMPARISON & Substitution & \makecell{Arg1-as-subst (Arg2-``head'');\\arg2-as-subst (Arg1-``head'')}
				
				\\[1ex] 
				\hline \hline
		\end{tabular}}
		\caption{\label{tab:asymmetric} 11 asymmetric PDTB annotations as the head or dependent in dependency parsing}
	\end{table}

	By contrast, the remaining 12 PDTB tag classes (we merely used 7 of 12 tags in dependency parsing) are taken as symmetric, that is, when two discourse units form a relation that belongs to these 12 classes, the two discourse units are treated as equally important, which is similar to mutiple-nuclear structure in the RST-DT. At this moment, we treat the second discourse unit as the ``head'' when the two discourse units form symmetric structure or multi-nuclear relation. Such annotations can be found in Chinese (TED-CDB, \citep{long2020ted}) and TED Multilingual Discourse Bank (TED-MDB, translated parallel corpus among English, Polish, German, Russian, European Portuguese, and Turkish) \citep{zeyrek2020ted}. Using the two strategies, we can successfully convert PDTB discourse relations into dependency representations. The following section discusses the conversion procedure in greater detail.

	\subsubsection{Example}
	
	We used a typical PDTB example to illustrate how to convert PDTB annotations in local dependency representations. The example, \texttt{``WSJ\_0618''} includes complicated structures and can represent the majority cases in the PDTB (also see \citet{xue2011multilingual}). To save the space, we put this complicated example and its annotations to the Appendix (A.1).
	
	\begin{figure}
		\centering	
		
		\includegraphics[width=\textwidth]{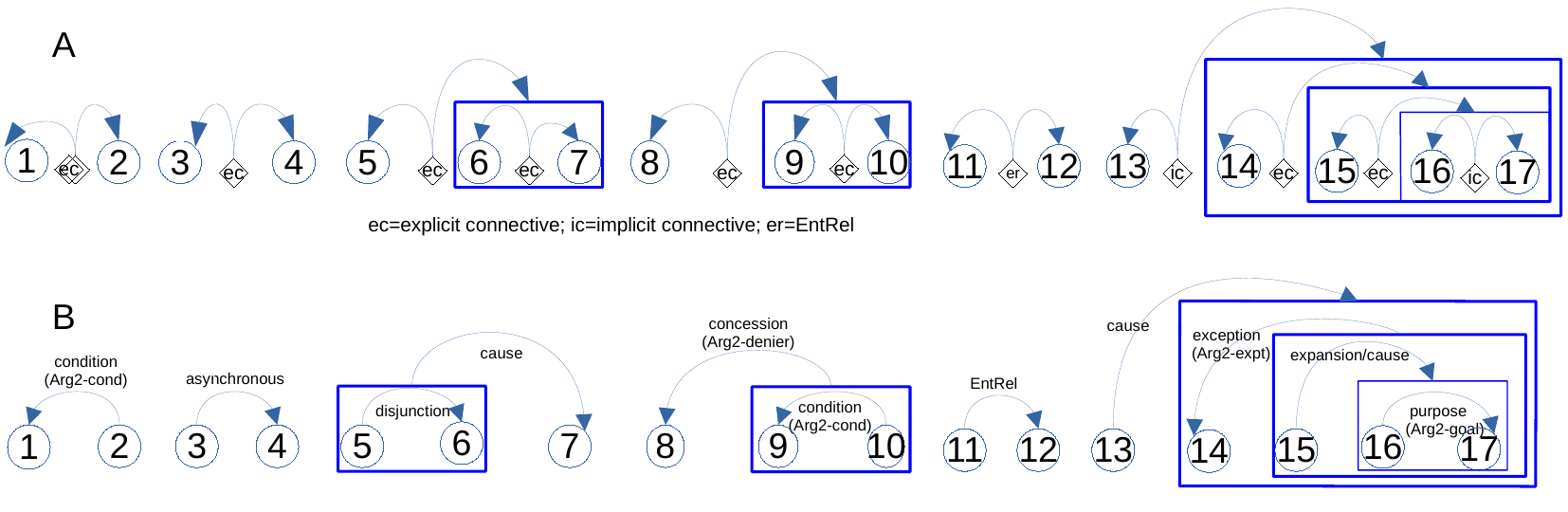}
		\caption{Panel A represents the original PDTB annotations (\texttt{WSJ\_0618}), that is, a discourse connective is a \textit{head} governing two discourse units. Panel B represents the converted data structure from Panel A.} 
		\label{fig:conv}
	\end{figure} 
	
	\begin{table}[h!]
		\centering
		\scalebox{0.82}{
			\begin{tabular}{|l|l|l|l|l|l|}
				\hline
				\makecell{	node 1 \\(dependent)} & \makecell{node 2 \\(head)} & \makecell{linear \\distance} & (PDTB) Sense & (PDTB) Class & (PDTB) Type \\ [0.5ex]
				\hline\hline 
				1 & 2 & 1 & EXPANSION & Condition  & Arg2-as-cond\\
				3 & 4 & 1 & TEMPORAL &  Asynchronous & Sucession\\
				5 & 6 & 1 & CONTINGENCY & Cause & Reason \\
				6 & 7 & 1 & EXPANSION &  Disjunction & NA \\
				9 & 8 & 1 & COMPARISON & Concession & Arg2-as-denier \\
				10 & 9 & 1 & COMPARISON & Condition & Arg2-as-cond \\
				11 & 12  & 1 & EntRel & NA & NA \\
				13 & 14  & 1 & CONTINGENCY & Cause & reason \\	
				17 & 14 & 3 & COMPARISON & Exception & Arg2-as-expect \\
				15 & 17 & 2 & CONTINGENCY & Cause & Result \\
				16 & 17  & 1 & COMPARISON & Purpose & Arg2-as-goal\\[1ex] 
				\hline\hline
		\end{tabular}}
		\caption{\label{tab:conversion}\texttt{WSJ\_0618} PDTB annotations are converted into dependency representations. The conversion result is shown in Table 2.}
	\end{table}
	
	
	PDTB3.0 annotations contain a large amount of information on asymmetric structure as shown in Table \ref{tab:asymmetric}. 
	For instance, in the example of texttt{WSJ\_0618}, node 9 (i.e., the 9$^{\text{th}}$ discourse unit) and node 10 (i.e., the 10$^{\text{th}}$ discourse unit) form condition relation with ``Arg2-as-cond''. This suggests that node 10 is a conditional clause and the node 9 is head. Further, the node 8 forms a concession relation (``Arg2-as-denier'') with the integrated unit construed by node 9 and node 10. This means that node 8 is head and the constituency unit construed by node 9 and node 10 is the subordinate. However, node 9 is the head for the combination of node 9 and node 10. This way, we can see that node 8 is the head and its subordinate is the node 9. We visualized all of these, shown in Fig. \ref{fig:conv}. 
	
	Still using the example of \texttt{WSJ\_0618}, node 16 and node 17 forms puprose relation (``Arg2-as-goal''), and node 17 is the head in this relationship. Note node 15 forms a cause (or expansion) relation with the combination of node 16 and node 17. As we know, ``cause'' is a symmetric type, and we therefore adopt the rule that the second node is the ``head'' when the nodes form a symmetric structure. Node 17 is the head in the constituent consisting of node 16 and node 17. This way, node 17 is the head and node 15 is dependent. More complicatedly, node 14 forms the exception relation. ``Arg2-as-excpt'' with the macro-structure composed by node 15, node 16, and node 17. In Table \ref{tab:asymmetric}, ``Arg1'' is the head but ``arg2'' is dependent, node 14 is the head, and node 17 is dependent. The conversion result is shown in Table \ref{tab:conversion}.  \footnote{The conversion code for depedency from PDTB is available at: \url{https://github.com/fivehills/discourse-corpora-resources}} 
	
	Ultimately, using the discourse distance equation, we can obtain the local discourse distance for this text: $(1+1+1+1+1+1+1+1+3+2+1)/11$$=1.27$. We will discuss the calculation of local discourse distance in the following section.

	
	\subsection{Discourse distance \& the variation of dependency distance}
	We can analyze discourse dependency data by applying dependency grammar algorithms. In a dependency relation, the linear distance between a head and dependent can potentially be utilized to provide a measure for assessing the depth in sentence processing (\citealp{hudson2007language}; \citealp{liu2008dependency}; \citealp{cong2014approaching}). This means that dependency parsing and dependency distance algorithms are helpful in quantitatively investigating the connection between discourse relations and discourse units. There is a linear distance that runs between any given ``head'' node and any given ``dependent'' node. We can obtain PDTB’s ``discourse distance'' from each linear depedency distance in a text. 

	The calculation of ``discourse distance'' in what follows uses the \textit{dependency distance} algorithm. \citet{liu2008dependency} used the term \textit{dependency distance}, and calculated the \textit{mean dependency distance} (MDD) of a sentence or a text, using the following one formula:
	\begin{equation}
		MMD(RST \ \ text)=\frac{1}{n-1}\sum_{i=1}^n|DD_i|
	\end{equation}
	In equation (1), \textsl{n} is the number of discourse units in the text and \textsl{DD{i}} is the dependency distance of the \textsl{i-th} dependency link of the text. 
	By contrast, PDTB local dependency does not contain the root, meaning that subtracting 1, is necessary, as shown in equation (2). With regard to calculating PDTB dependency distance, the number of discourse units is based on the actual number of participants as heads or dependents rather than the number of all discourse units in a text. For instance, in terms of PDTB annotations, \texttt{WSJ\_0618} includes 11 discourse units which are assigned with heads or dependents. The number of discourse units involving discourse dependency is 11. Computing PDTB dependency distance is shown in equation (2).
	\begin{equation}
		MMD(PDTB \ \ text)=\frac{1}{real\ \ n}\sum_{i=1}^n|DD_i|
	\end{equation}
	
	The ``discourse distance'' of global dependencies in Table \ref{table:1} (the top panel) is ``3.1''. The ``discourse distance'' of local dependencies in Table \ref{table:1} (the bottom panel) is ``1.13''. Moreover, we introduce \textbf{standard variation (SD)} of the data on dependency distances for a text because SD can tell how dispersed the dependency distances are. A small or low SD would indicate that many of the dependency distances are clustered tightly around the mean and tend to be processed easily by humans. 
	Using SD in statistics, we can calculate the SD of dependency distances for the top panel in Table \ref{fig:conv}, which is ``2.28'' and the SD of dependency distances at the bottom panel in Table \ref{table:1} is ``0.35''.
	
\begin{figure}
		\centering
		\includegraphics[width=\columnwidth]{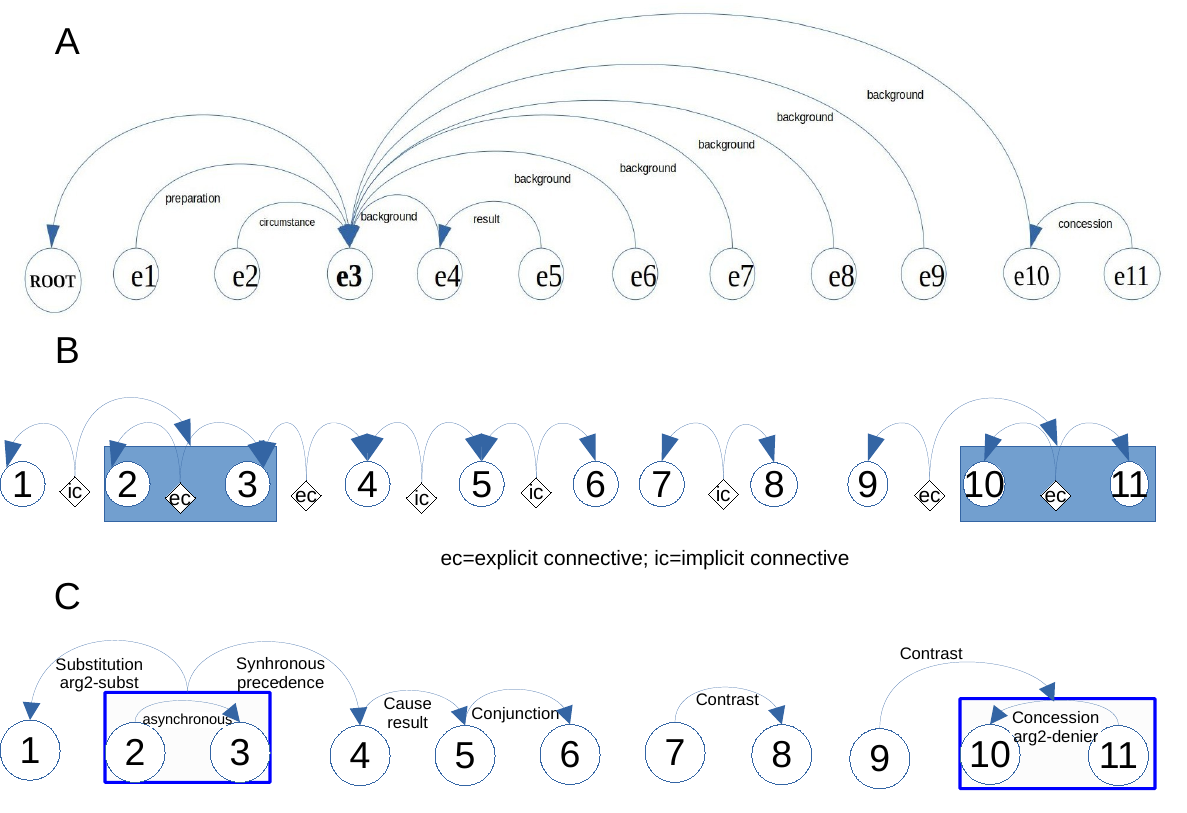}
		\caption{\underline{Panel A:} The RST tree of Fig.\ref{fig:1} converts to a dependency tree. An EDU in RST is similar to a word in syntactic dependency analysis. The node where the arrow point is located is a \textit{head (or governor)}. Here, ``e'' denotes ``EDU''. \underline{Panel B:} The text in Fig.\ref{fig:1} can be annotated by the styles of PDTB. Given the EDUs in this text remain the same in PDTB annotations, the connectives are heads that form dependencies. \underline{Panel C:} Such relations in B can be treated as dependencies beteeen EDUs, as well as co-occurring network data, as shown in the bottom panel of Table \ref{table:1}.}
		\label{fig:2}
		
	\end{figure}

	Another example (\textbf{Fig. \ref{fig:1}}) illustrates how annotations for the same text by RST and PDTB can be converted into global dependency and local dependency respectively. \citet{hirao2013single} and \citet{li2014text} both developed algorithms for converting RST discourse representations into dependency structures. Their discourse dependency framework is adopted from a syntactic dependency with words replaced by EDUs. Fig.\ref{fig:2} (Panel A) illustrates how the RST tree of Fig.\ref{fig:1} is converted into a dependency tree by adopting the method of \citet{hirao2013single}. In Fig.\ref{fig:2} (Panel A), binary discourse relations are represented by dominant EDU (``head/governor'') to subordinate EDU (``dependent''). Table \ref{table:1} (Top Panel) shows that the dependency representations from Fig.\ref{fig:2} (Panel A) can be treated as network data. Dependencies reflect the global and rhetorical relations in RST.
	Although a dependency that is based on a discourse connective as head in PDTB is different to the RST relation dependency, such a dependency in PDTB can be treated as the relation dependency between two EDUs in practice because people tend to be concerned with the relation between two EDUs in the PDTB. Panel B in Fig. \ref{fig:2} presents the PDTB annotation for Fig. \ref{fig:1}. We can convert Panel B into Panel C, as shown in Fig.\ref{fig:2}. In Panel C of Fig. \ref{fig:2}, the EDUs in PDTB are connected with each other, and they can be treated as (local) dependencies.
	
	Following Equation (2), the ``discourse distance'' of global dependencies in Table \ref{table:1} (the top panel) is (2+1+1+1+3+4+5+6+7+1)/(11-1) = ``3.1''. The data on the dependency containing ``0'' in the nucleus (i.e., a ROOT relation) is not computed in our algorithm. The ``discourse distance'' of local dependencies in Table \ref{table:1} (the bottom panel) is (2+1+1+1+1+1+1+1)/(8) = ``1.13'' (the real number of discourse units involving in dependency is 8).  When there are a number of texts, we can obtain the average discourse distance for these texts through calculating the mean of all values of discourse distance in these texts.

	\begin{table}
		
		\centering
		\scalebox{0.78}{
			\begin{tabular}{||l|l|l|l|l|l||}
				\hline
				\makecell{satellite\\(dependent)} & \makecell{nucleus\\(head/governor)}& \makecell{dependency\\ distance} & frequency & (RST) relation & (RST) class \\ [0.5ex] 
				\hline\hline 
				1 & (2-11)3 & 2 & 1 & preparation & ELABORATION\\
				2 & (3-11)3 & 1& 1 &  circumstance & BACKGROUND\\
				3 & 0 & NA & 1& ROOT & NONE\\
				4 & 3 & 1& 1 &  background & BACKGROUND \\
				5 & 4 & 1 & 1 & result & CAUSE \\
				6 & (3-5)3 & 3 & 1 & background & BACKGROUND \\
				7 & (3-6)3 & 4 & 1 & background & BACKGROUND \\
				8 & (3-6)3  & 5 & 1 & background & BACKGROUND \\
				9 & (3-8)3 & 6 & 1 & background & BACKGROUND \\
				10 & (9-11)3 & 7 & 1 & background & BACKGROUND \\
				11 & 10 & 1 & 1 & concession & CONTRAST \\	[1ex] 
				\hline\hline
				\centering
				\makecell{	node 1 \\(dependent)} & \makecell{node 2 \\(head)} & distance & (PDTB) Sense & (PDTB) Class & (PDTB) Type \\ [0.5ex]
				\hline\hline 
				3 & 1 & 2 & EXPANSION & substitution   & Arg2-as-subst\\
				2 & 3 & 1 & TEMPORAL &  Asynchronous & NA\\
				3 & 4 & 1 & TEMPORAL & Synhronous & precedence \\
				4 & 5 & 1 & CONTINGENCY &  Cause & result \\
				5 & 6 & 1 & EXPANSION & Conjunction & NA \\
				7 & 8 & 1 & COMPARISON & Contrast & NA \\
				9 & 10 & 1 & COMPARISON & Contrast & NA \\
				11 & 10  & 1 & COMPARISON & Concession & arg2-as-denier \\	[1ex] 
				\hline\hline
		\end{tabular}}
		
		\captionof{table}{RST discourse (global) dependency representations (top) and PDTB (local) dependenciy representation (bottom) in the example of Fig. \ref{fig:1}. Two kinds of dependencies (the top and the bottom) can be taken as network data.}
		\label{table:1}
	\end{table}

	We converted all WSJ texts with both RST and PDTB annotations (375 texts) into global dependencies (from RST) and local dependencies (from PDTB) respectively by using the aforementioned methods. The following section will use two types of dependency data to carry out experiments in order to verify their validity.

	\section{Experiments}

	To expand on the evaluation and validation of the converted dependency data, we employed several modified SOTA discourse parsers to test the validity of the dependency representations derived from PDTB-style corpora. The experimental setup involved using datasets from English, Chinese, and several other languages for training and testing. We applied specific pre-processing steps to prepare the data and tasked the discourse parsers with relation identification and argument extraction. The evaluation metrics included F1 score, precision and recall, exact match accuracy, and labeled attachment score (LAS) for dependency parsing. Our cross-linguistic validation extended beyond English and Chinese, addressing any challenges or adaptations required for different languages and providing comparative results across them. The following details how we implemented these experiments and validated the converted data in two sections.

	\subsection{Parser evaluations on PDTB dependency representations across languages}
	The methods adopted were tested experimentally by using the local discourse dependency data converted from the PDTB. In order to make a comparison with the RST-DT texts, we experimented on the 375 WSJ texts with both RST and PDTB annotations. The 375 WSJ texts with PDTB annotations were converted into local dependency representations. The training part of the corpus is composed of 303 texts, while the test part consists of 36 texts and 36 texts were respectively taken as the development part and test part. Meanwhile, similar methods were applied in the different PDTB- style corpora of other languages (Chinese, TED-CDB, \citet{long2020ted}; TED-MDB, \citet{zeyrek2020ted}), making PDTB relations into become local discourse dependencies. However, these corpora have not corresponding RST ones.
	
	Two discourse dependency parsers were refined to parse the PDTB dependency data in the present study. \citet{nishida2022out} modified two state-of-art discourse dependency parsers. The first parser is Arc-Factored Model which combines the BERT-based biaffine attention model \citep{dozat2016deep} and Hierarchical Eisner Algorithm \citep{zhang2021adapting} (abbreived as ``NISHIDA22-ARC-MOD''). The second parser, Stack-pointer discourse dependency parser, uses a BERT-based pointer network  (``NISHIDA22-STP-MOD''). After modifications and refinements, the two discourse parsers focus solely on parsing discourse dependency relations, but without considering the dependency tree task. For instance, unlike traditional dependency or RST parsing tasks, we need not to build an RST tree, which makes our parsing tasks easier. The number of PDTB dependency relations tags is 19 (i.e., 18 second-level tags + EntRel). 7 tags are not required to recognize the difference between head or dependent (i.e., ``synchronous'', ``asynchronous'', ``cause'', ``contrast'', ``similarity'', ``conjunction'', ``disjunction''), but the remaining 10 are required (shown in Table \ref{tab:asymmetric}). We modified the two parsers to parse the PDTB dependency data to account for this.
	
	The following details some \textbf{hyper-parameters} we applied in using the refined dependency parsers. When parsing English dependency data, we employed specific hyperparameters as follows: the dimensionality of MLPs within the (ARC) and the shift-reduce model (STR) were set at 120 and 150, respectively. To optimize the model, we used AdamW and Adam optimizers. These optimizers were directed towards refining the transformer's parameters ($\theta_{\text{bert}}$) and the task-specific parameters ($\theta_{\text{task}}$), respectively, aligning with the methodology of SpanBERT \citep{joshi2020spanbert}. In the initial phase, we trained the foundational models on the labeled source dataset with the following hyperparameters: batch size of 1, a learning rate (lr) of $2 \times 10^{-5}$ for $\theta_{\text{bert}}$, lr of $1 \times 10^{-4}$ for $\theta_{\text{task}}$, and 3,000 warmup steps. Subsequently, we implemented a singular bootstrapping approach, specifically co-training, but without utilizing other bootstrapping methods. During this process, the model's hyperparameters were configured as follows: batch size of 1, lr of $2 \times 10^{-6}$ for $\theta_{\text{bert}}$, lr of $1 \times 10^{-5}$ for $\theta_{\text{task}}$, and 6,000 warmup steps. The training duration for all methodologies extended to a maximum of 30 epochs. We also integrated an early stopping mechanism, suspending training when validation LAS exhibited no improvement for a consecutive span of 10 epochs.

	We adopted a similar strategy for parsing the Chinese dependency data. The dimensionality of MLPs within the arc-factored model (ARC) and the shift-reduce model (STR) remained consistent, being set at 80 and 95, respectively. The optimization approach employed AdamW and Adam optimizers. These were employed to refine the parameters of the Chinese BERT model ($\theta_{\text{Chinese-BERT}}$) and the task-specific parameters ($\theta_{\text{task}}$). Specifically, we configured the hyperparameters as follows: batch size of 1, a learning rate (lr) of $2 \times 10^{-4}$ for $\theta_{\text{Chinese-BERT}}$, lr of $1 \times 10^{-3}$ for $\theta_{\text{task}}$, and 2000 warmup steps. Similar to the English case, a single bootstrapping approach (co-training) was applied, excluding alternative methods. During this phase, the model's hyperparameters were set as follows: batch size of 1, lr of $2 \times 10^{-5}$ for $\theta_{\text{Chinese-BERT}}$, lr of $1 \times 10^{-3}$ for $\theta_{\text{task}}$, and 5,000 warmup steps. Training was conducted for a maximum of 30 epochs and early stopping was enacted when validation LAS did not exhibit any improvement for a consecutive span of 10 epochs. Moreover, our strategies involved the utilization of ``multilingual BERT'' \citep{devlin2018bert} in relation to the data dependencies of the MDB in six languages. Due to the limited scale of the dataset in each language, we made the necessary adjustments to the hyperparameters accordingly. Note that the final results for all six languages were derived from the mean of their respective outcomes.

	\begin{table}
		\centering
		\scalebox{0.96}{
			\begin{tabular}{|l|l|l|}
				\hline
				Data & \scriptsize{NISHIDA22-ARC-MOD} & \scriptsize{NISHIDA22-STP-MOD} \\
				\hline\hline
				PDTB-dependency dev. (English)  & 65.7\% & 66.4\% \\
				
				PDTB-dependency test (English)  & 65.8\% & 66.5\% \\
				RST-DT dependency (English) & 62.9\% & 62.3\% \\
				\hline
				TED-CDB depedency (Chinese) & 57.2\% & 58.1\% \\
				\hline
				TED-MDB dependency (six languages) & 56.2\% & 55.7\% \\
				\hline
		\end{tabular}}
		\caption{Unsupervised discourse dependency parsing results on the PDTB dependency data and RSD-DT dependency data. The evaluation metric is the Unlabeled Attachment Score (UAS). We did not report the results of Labeled Attachment Score (LAS) because LAS scores are lower than UAS ones on average.  Note that RST-DT parsing tasks were not done in the present study.}
		\label{compare}
	\end{table} 
	
	After using the refined parsers on the converted data, we report the result using the \texttt{micro-average F1 score}, as shown in Table \ref{compare}. This result shows that two SOTA refined dependency discourse parsers are able to analyze the relations in PDTB dependency data. The same two parsers have performed a little better in analyzing PDTB dependency relations as compared with the performance on RST data. The reason for this is that PDTB dependency data do not require forming a tree structure and they just need to construe local dependencies, thus making the parsing tasks become simpler. Considering the stable performance by the two parsers, we have evidence supporting these thesis that PDTB dependency data can be automatically analyzed and that this will be useful in computational analysis.
	
	\subsection{The correlation between mean/SD discourse distance of RST and PDTB}

	At present, only English texts have both RST and PDTB3.0 annotations, which means correlation analyses are restricted to this language alone. We used the same 375 \texttt{WSJ} texts with both RST and PDTB annotations (English) to extract their global dependency and the local dependency respectively. Note that we applied the algorithms of \citet{hirao2013single} and \citet{li2014text} to extract RST dependencies respectively. After that, we applied the discourse distance algorithms to calculate the indexes of discourse distance for the global dependency (two types, \citet{hirao2013single} and \citet{li2014text}) and local dependency for each text. After obtaining the discourse distance for the global dependency and local dependency of each text, we used \textit{Pearson's correlation} to test the relationship between discourse distance of global dependency and that of local dependency for these 375 texts. 
	
	The result shows that the correlation between mean discourse distance of global dependency by \citet{hirao2013single} and that of local dependency reaches \textbf{82.69\%}(\textit{p}-value < 1.26e-12, df = 374), and the correlation between discourse distance of global dependency by \citet{li2014text} and that of local dependency reaches \textbf{79.23\%}(\textit{p}-value < 1.26e-12, df = 374). The correlation of SD of discourse distance and global dependency is \textbf{81.26\%} (\textit{p}-value < 1.57e-10, d f= 374). These correlation values show that two types of discourse distance are closely corelated. This result is basically consistent with the finding from \citet{demberg2017compatible} that there is 76\% mapping between RST and PDTB relations. Further, the correlation result indicates that the PDTB dependency data is valid and can be used for different types of computations. It also suggests that both types of discourse distance can be used for measuring textual complexity and quantifying other linguistic explorations (\citealp{davoodi2017contribution}; \citealp{sun2019computational}).

	The quantitative results from the parsing experiments are comprehensive, presenting performance scores for different relation types and comparing parser performance on original PDTB annotations versus converted dependency structures. We conducted statistical significance tests to validate any observed improvements. An in-depth \texttt{error analysis} revealed common challenges encountered during the parsing experiments, providing insights into areas where the conversion process might need refinement. We also compared the results of parsing the converted dependency data to a baseline of parsing the original PDTB annotations, demonstrating the value of the conversion process. Additionally, we discussed observations regarding the scalability and computational efficiency of parsing the converted dependency structures compared to the original annotations. These detailed evaluations and validations strengthen our claims about the validity and usefulness of the converted dependency data for discourse analysis across different frameworks and languages.

	
	\section{Discussion}

	As discussed by \citet{hayashi2016empirical} and \citet{morey2018dependency}, some information is lost in dependency conversions for RST discourse trees. However, the most important information can be retained. PDTB-dependency conversion seems to lose much less than the conversion between RST and dependency. The primary loss is the non-inclusion of information on discourse connectives. The other potential loss is that after some local constituency structure becomes dependencies, the inner connection is lost, as was discussed in the RST constituency conversion. However, PDTB-dependency conversion also has great benefits. For example, the implicit information on the head and dependent become more explicit in PDTB dependency representation. PDTB dependency representations enable additional computational possibilities.
	
	The present study reports the success of conversion for dependency from PDTB and this supports the validity of the PDTB dependency data. We claim that the dependency format can be successfully derived from RST, SRDT and PDTB corpora. However, we must be aware of the fact that the three types of dependency representations are still different. RST dependencies represent global connections and form a global dependency tree. SRDT dependencies can still basically represent global coherence, but they cannot form a dependency tree. PDTB dependencies just represent local coherence. Despite this, the three types of dependencies are closely related with each other, and can complement each other. For instance, global (RST) discourse dependencies and local (PDTB) discourse dependencies could work together to better characterize the discourse structure. The core of \texttt{Uinversal Dependeices} (UD) lies in its utilization of syntactic dependency relations, while also incorporating a substantial amount of morphological and syntactic information. This contrasts with traditional dependency grammar and dependency corpora (\citealp{nivre2016universal}; \citealp{de2021universal}). We expect that discourse dependencies in mutiple languages could play a similar role as UD. When these discourse corpora annotations are converted into dependency representations the possible benefits are unlimited (see the Appendix A.4).

	The use of dependency structures offers several key advantages for unifying different discourse frameworks and enabling more consistent analysis across corpora. Dependency representations provide a common format that can be applied across various discourse theories while preserving the original information from each framework. They offer flexibility in handling both simple and complex discourse relations, enabling more consistent quantitative analysis through metrics like "dependency distance." Computationally, dependency parsing leverages existing algorithms and tools from syntactic analysis. The approach has demonstrated cross-linguistic applicability, facilitating consistent discourse analysis across multiple languages. By converting different frameworks into a common dependency representation, researchers can more easily compare annotations from different theoretical perspectives, potentially bridging gaps between theories. Additionally, unified dependency representations can provide larger, more consistent datasets for training machine learning models in discourse parsing and analysis. Moreover, this dependency-based discourse structure framework can serve as a form of prompting for SOTA Large Language Models (LLMs). By incorporating this framework, LLMs can enhance their comprehension of discourse and textual structures, potentially leading to improved performance across various NLP tasks.  Overall, this approach allows researchers to overcome challenges posed by multiple discourse theories and annotation schemes, opening up new possibilities for computational discourse analysis and cross-framework studies.

	While our study demonstrates the potential of the dependency framework, there are limitations that warrant further investigation. The conversion process may not capture all nuances of the original annotations, particularly for complex discourse phenomena, and future work should focus on refining the conversion algorithms to address these limitations. Additionally, the current study focused primarily on PDTB and RST, and extending the framework to incorporate other discourse theories and annotation schemes would further validate its universality. More extensive cross-linguistic studies are also needed to fully explore the framework's applicability across a wider range of languages and discourse types. Future research directions could include developing new discourse parsing algorithms that directly employ the unified dependency representation, investigating the relationship between discourse dependency structures and other linguistic phenomena such as coreference or lexical cohesion, and exploring the application of the framework to discourse-level tasks in NLP, such as text summarization or coherence evaluation.

\section{Conclusion}

This study introduces a groundbreaking dependency framework that unifies discourse data analysis across diverse theoretical approaches. Our key findings demonstrate the successful conversion of PDTB annotations into dependency structures, preserving original information while enabling more unified analysis. The validity of this converted data has been confirmed through extensive testing using state-of-the-art discourse parsers across multiple languages. Notably, we discovered a strong correlation between RST and PDTB dependencies, suggesting underlying structural similarities between these frameworks. Furthermore, the cross-linguistic applicability of our dependency approach has been validated across English, Chinese, and several other languages, underscoring its versatility and robustness.

The significance of this work for discourse analysis is substantial, providing a unified framework that overcomes previous limitations in comparing and integrating diverse discourse corpora. This dependency representation enables more consistent quantitative analysis of discourse relations, opening new avenues for computational discourse studies. By bridging different discourse frameworks, our approach facilitates a more comprehensive understanding of discourse structure, potentially leading to new theoretical insights. Additionally, the unified dependency format creates larger, more consistent datasets for training machine learning models, potentially advancing automated discourse parsing and analysis. The cross-linguistic validity of this framework enhances its potential for comparative discourse studies across languages, laying the groundwork for future advancements in both theoretical and computational approaches to discourse analysis. In short, this novel dependency framework represents a significant step forward, offering a powerful tool for researchers to explore discourse structure more comprehensively and consistently across different theoretical perspectives and languages, promising to deepen our understanding of how language creates meaning at the discourse level.

	\bibliographystyle{apalike}
	\bibliography{custom1.bib}
	
	\appendix
	\onecolumn
	\section{Appendix}
	\label{sec:appendix}

	\subsection{PDTB Example and annotations}
	
	The following is the original text of  \texttt{``WSJ\_0618''}. Note that the ordinal numbers are added following the PDTB annotations.

	\textit{The head of the nation's largest car-dealers group is telling dealers to "just say no"(1) when auto makers pressure them to stockpile cars on their lots(2).} 
	
	\textit{In an open letter that will run today in the trade journal Automotive News, Ron Tonkin, president of the National Car Dealers Association, says dealers should cut their inventories to no more than half the level traditionally considered desirable.} 
	
	\textit{Mr. Tonkin, who has been feuding with the Big Three (3) since he took office earlier this year(4), said that with half of the nation's dealers losing money (5) or breaking event (6) it was time for "emergency action."(7)} 
	
	\textit{U.S. car dealers had an average of 59 days' supply of cars in their lots at the end of September, according to Ward's Automotive Reports (8).
		But Mr. Tonkin said dealers should slash stocks to between 15 and 30 days (9) to reduce the costs of financing inventory(10).} 
	
	\textit{His message is getting a chilly reception in Detroit, where the Big Three auto makers are already being forced to close plants because of soft sales and reduced dealer orders (11). Even before Mr. Tonkin's broadside, some large dealers said they were cutting inventories.(12)} 
	
	\textit{Ford Motor Co. and Chrysler Corp. representatives criticized Mr. Tonkin's plan as unworkable (13). It "is going to sound neat to the dealer (14) except when his 15-day car supply doesn't include the bright red one (15) that the lady wants to buy (16) and she goes up the street to buy one,"(17) a Chrysler spokesman said.}
	
	\hspace{1em}\\
	
	The following are the PDTB annotations for  \texttt{``WSJ\_0618''}.
	
	\noindent\texttt{\tiny{Explicit|96..100|||||9..78|when|Contingency.Condition.Arg2-as-cond||||||79..94||||||101..158|||||||||||96..100|PDTB2::wsj\_0618::96..100::SAME|}}
	\texttt{\tiny{
			Explicit|464..469||||||since|Temporal.Asynchronous.Succession||||||424..463||||||470..502|||||||||||464..469|PDTB2::wsj\_0618::464..469::SAME|}}
	\texttt{\tiny{
			Explicit|514..518||||||with|Contingency.Cause.Reason||||||509..513;579..612||||||519..577|||||||||ARGM-ADV|be|514..518|PDTB3|}}
	\texttt{\tiny{
			Explicit|561..563||||||or|Expansion.Disjunction||||||548..560||||||564..577|||||||||||561..563|PDTB3|}}
	\texttt{\tiny{
			Explicit|755..758|||||759..774|but|Comparison.Concession.Arg2-as-denier||||||617..713|||||775..871|||||||||||755..758|PDTB2::wsj\_0618::755..758::CHANGED|}}
	
	\noindent\texttt{\tiny{Implicit||||||759..774|if they are|Contingency.Condition.Arg2-as-cond||thereby|Expansion.Manner.Arg1-as-manner|||775..828||||||829..871|||||||||ARGM-PRP|slash|829|PDTB3|}}
	\texttt{\tiny{
			EntRel||||||||||||||875..1049||||||1051..1140|||||||||||1051|PDTB3|}}
	\texttt{\tiny{
			Implicit|||||||because|Contingency.Cause.Reason||||||1205..1236|||||1144..1204|1238..1412|||||1415..1440||||||1238|PDTB2::wsj\_0618::1238::SAME|}}
	\texttt{\tiny{
			Explicit|1279..1285|||||1415..1440|except|Expansion.Exception.Arg2-as-excpt||||||1238..1278||||||1286..1412|||||||||||1279..1285|PDTB2::wsj\_0618::1279..1290::CHANGED|}}
	\texttt{\tiny{
			Implicit|||||||so|Contingency.Cause.Result||||||1291..1374||||||1375..1412|||||||||||1375|PDTB3|LINK1}}
	\texttt{\tiny{
			Explicit|1375..1378|||||1415..1440|and|Expansion.Conjunction||||||1291..1374||||||1379..1412|||||||||||1375..1378|PDTB2::wsj\_0618::1375..1378::CHANGED|LINK1}}
	\texttt{\tiny{
			Implicit||||||1415..1440|in order|Contingency.Purpose.Arg2-as-goal||||||1375..1401||||||1402..1412|||||||||ARGM-PRP|go|1402|PDTB3|}}

	\subsection{The PDTB annotation system}
	
	\begin{figure*}
		\includegraphics[width=0.96\textwidth]{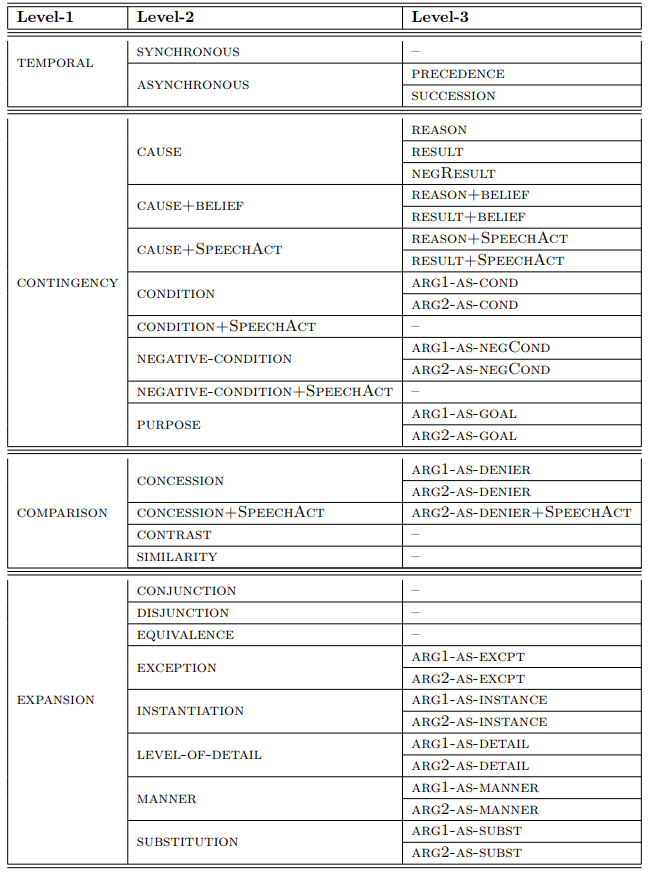}
		\caption{PDTB-3 Sense Hierarchy. The leftmost column contains the Level-1 senses and the middle column, the Level-2 senses.}
			\label{fig:pdtbann}	
	\end{figure*}
	

	\subsection{RST, PDTB, SRDT and discourse dependency corpora in different languages (Tables 5-8)}\footnote{More information on these discourse corpora can be seen at \url{https://github.com/fivehills/discourse-corpora-resources/blob/main/discourse.md}}
	\begin{table}
		\setlength{\tabcolsep}{2.5pt}
		\centering
		\caption{RST-style multilingual corpora}
		\resizebox{\textwidth}{!}{
			\begin{tabular}{||c c c c||}
				\hline
				\textbf{language} & \textbf{corpus name} & \textbf{size} & \textbf{genre} \\ [0.5ex]
				\hline\hline
				Basque & RST Basque Treebank & 15.5K words & abstracts \\
				\hline 
				Chinese & Macro Chinese Discourse Treebank & 720 texts & newspaper \\
				\hline
				Chinese & Chinese/Spanish Treebank & 100 texts & different genres \\
				\hline
				Chinese & `Run-on' sentences & 500 sentence & fiction \\
				\hline
				Dutch & Dutch RUG Corpus & 6K words & several written genres \\
				\hline
				English & RST Discourse Treebank & 176K words & newspaper \\
				\hline
				English & GUM & 176K words & different genres \\
				\hline
				English &  SciDTB  &  1.3K texts & scientific abstracts \\
				\hline
				German & Potsdam Commentary Corpus & 44K words & newspaper \\
				\hline
				Portuguese & CST News & 47K words & newspaper\\
				\hline
				Spanish &  RST Spanish Treebank & 52K words & written science \\ [0.5ex]
				\hline
			\end{tabular}
		}
		
		\bigskip
		
		\caption{PDTB-style multilingual corpora}
		\resizebox{\textwidth}{!}{
			\begin{tabular}{||c c c c||}
				\hline
				\textbf{language} & \textbf{corpus name} & \textbf{size} & \textbf{genre} \\[0.5ex]
				\hline\hline
				Chinese & Chinese Discourse Treebank & 70K words & newspaper\\
				\hline
				Chinese & TED-CDB & 72 texts & talks\\
				\hline 
				Czech & Prague Discourse Treebank & 50K sentences & newspaper \\
				\hline
				English & Penn Discourse Treebank & one million words & newspaper \\
				\hline
				English & TED-MDB & 6 texts & TED talks \\
				\hline
				English & BioDRB & 112K words & Biomedical articles\\
				\hline
				French & DisFrEn & 16K words & spoken genres \\
				\hline
				German & Potsdam Commentary Corpus & 44K words & newspaper \\
				\hline
				\makecell{German, Polish, Russian,\\ Portuguese, and Turkish\\(parallel corpora by translation)}  & TED-MDB & 6 texts & TED talks \\
				\hline
				Hindi & Hindi Discourse Relation Bank & 400K words & newspaper \\
				\hline
				Italian & Luna Corpus & 25K words & dialogue \\
				\hline
				Modern Standard Arabic & Leeds Arabic DTB & 166K words & newspaper\\
				\hline
				Iranian & Persian Discourse Treebank & 30,000 sentences & NA\\
				\hline
				Turkish & METU-TDB Corpus & 500K words & different genres \\[0.5ex]
				\hline
			\end{tabular}
		}
		
		\bigskip
		
		\caption{SDRT-style multilingual corpora}
		\resizebox{\textwidth}{!}{
			\begin{tabular}{||c c c c||}
				\hline
				\textbf{language} & \textbf{corpus name} & \textbf{size} & \textbf{genre} \\[0.5ex]
				\hline\hline
				English &  STAC & 1081 dialogues & multiparty dialogues \\
				\hline
				English & Molweni & 10K dialogues & multiparty dialogues \\
				\hline
				English & Parallel Meaning Bank & 10 documents & NA\\
				\hline
				French & ANNODIS & NA & NA\\
				\hline
				\makecell{German, Dutch \\ and Italian \\(parallel corpora by translation)} & Parallel Meaning Bank & 10 documents in each language & spoken genres \\[0.5ex]
				\hline
			\end{tabular}
		}
		
		\bigskip
		
		\caption{Discourse dependency-style multilingual corpora}
		\resizebox{\textwidth}{!}{
			\begin{tabular}{||c c c c||}
				\hline
				\textbf{language} & \textbf{corpus name} & \textbf{size} & \textbf{genre} \\[0.5ex]
				\hline\hline
				Chinese & UnifiedDep & 2793 documents & newspaper \& academic \\
				\hline 
				English & SciDTB & 1355 texts & ACL paper abstracts \\
				\hline
				English & Molweni & 10K dialogues & multiparty dialogues \\
				\hline
				English & COVID19-DTB & 300 documents & medical articles\\[0.5ex]
				\hline
			\end{tabular}
		}
	\end{table}


\subsection{Applications of discourse dependencies}

The following discusses the benefits of this approach. \textbf{1)} A unified method can be applied to extract discourse corpora, which is the objective of the current study. \textbf{2)} Converting data into dependency format allows for the adoption of diverse dependency analysis algorithms, enabling a deeper exploration of important issues such as textual complexity, linear features in discourse structure, and language efficiency. Dependency data overcomes the limitations set by original corpora structures and offers a more extensive algorithmic support, fostering connections with various fields in linguistics and computational linguistics. Dependency algorithms can help in exploring the following important issues: textual complexity, linear features in discourse structure and language efficiency etc. \textbf{3)} Discourse dependency representations derived from these discourse corpora can be seen as network data. By providing a visual representation of the network data, we can better observe the global/local topological connection of discourse units. \textbf{4)} The dependency algorithms make it possible to do typological language investigations at the textual level (see discourse corpora resources in different language shown in Tables 5, 6, 7 \& 8 in the Appendix A.3). The role of discourse dependency is expected to be similar to that of syntactic dependencies which allow exploration using a number of algorithms. Additionally, the data on multilingual discourse dependencies could be similar to \textit{UD}. We expect that the multilingual discourse data can play a similar role to UD for linguistic research and computational linguistics. We also expect that multilingual discourse dependency data can contribute to language typology studies and discourse semantic parsing, like the role of UD (\citealp{levshina2019token}; \citealp{hahn2021modeling}; \citealp{mcdonald2013universal}; \citealp{qi2018universal}). \textbf{5)} Studies of syntactic dependency have achieved great successes in theoretical linguistics, computational linguistics, cross-lingual studies, and cognitive studies (\citealp{gibson2019efficiency}; \citealp{temperley2018minimizing}; \citealp{phillips2005erp}). Dependencies at the textual level could draw on and adapt research from and stand in contrast to the studies on syntactic dependencies. In this way, the research related to discourse dependency can unify different levels of linguistics and different areas of language sciences and it has the potential to make further contributions to the language sciences.

\end{document}